\documentclass[10pt,twocolumn,letterpaper]{article}

\usepackage{cvpr}              

\definecolor{cvprblue}{rgb}{0.21,0.49,0.74}
\usepackage[pagebackref,breaklinks,colorlinks,allcolors=cvprblue]{hyperref}
\usepackage{multirow}
\usepackage{graphicx}
\usepackage{svg}  
\usepackage{subcaption}  

\def\paperID{12779} 
\def\confName{CVPR}
\def\confYear{2025}

\title{Revealing Subtle Phenotypes in Small Microscopy Datasets Using Latent Diffusion Models}

\author{
    Anis Bourou$^{1,2}$\footnotemark[1] \quad Biel Castaño Segade$^{1}$\footnotemark[1] \quad Thomas Boyer$^{1}$
    \quad Valérie Mezger$^{2}$\\
     \quad Auguste Genovesio$^{1}$ \vspace{0.3em} \\
    {\normalsize $^1$Ecole Normale Supérieure} \\
    {\normalsize $^2$Université Paris Cité} \\
    {\normalsize \texttt{Correspondance:auguste.genovesio@ens.psl.eu}}\\
    {\normalsize $^*$Equal Contribution}
}

\begin{document}
\maketitle
\begin{abstract}
Identifying subtle phenotypic variations in cellular images is critical for advancing biological research and accelerating drug discovery. These variations are often masked by the inherent cellular heterogeneity, making it challenging to distinguish differences between experimental conditions. Recent advancements in deep generative models have demonstrated significant potential for revealing these nuanced phenotypes through image translation, opening new frontiers in cellular and molecular biology as well as the identification of novel biomarkers. Among these generative models, diffusion models stand out for their ability to produce high-quality, realistic images. However, training diffusion models typically requires large datasets and substantial computational resources, both of which can be limited in biological research. In this work, we propose a novel approach that leverages pre-trained latent diffusion models to uncover subtle phenotypic changes. We validate our approach qualitatively and quantitatively on several small datasets of microscopy images. Our findings reveal that our approach enables effective detection of phenotypic variations, capturing both visually apparent and imperceptible differences. Ultimately, our results highlight the promising potential of this approach for phenotype detection, especially in contexts constrained by limited data and computational capacity.
\end{abstract}    
\section{Introduction}
\label{sec:intro}

In recent years, generative models have undergone rapid and accelerating advancements~\cite{ddim,ddpm,normalizing_flow,generative_models_survey,bourou_conditional}, resulting in their widespread adoption across a variety of fields. Notably, these models have made significant contributions to biological research. For example, they have been employed in protein design~\cite{protein_design}, predicting protein structures~\cite{alphafold}, integrating cancer data~\cite{cancer_data}, synthesizing biomedical images~\cite{biomedical_1,biomedical_2}, predicting molecular structures~\cite{molgan,denovo}, and identifying phenotypic cell variations~\cite{bourou_1,bourou_2,Lamiable2023}.

Identifying phenotypic variations in biological images is crucial for advancing our understanding of biological processes. Detecting these differences can be particularly challenging due to the high degree of biological variability, yet it holds immense potential for enhancing disease understanding, discovering novel biomarkers, and developing new therapeutics and diagnostics~\cite{cellular_profiling,cellular_profiling_2,cellular_profiling_3}. Traditional methods for identifying these phenotypes often rely on cell segmentation and the quantification of features such as intensity, shape, and texture~\cite{cellular_profiling}. Recently, deep learning techniques, particularly generative models~\cite{Lamiable2023,bourou_1,bourou_2}, have been applied to automate and refine this process, enabling the identification of more interpretable and biologically meaningful features. Among these approaches, diffusion models have emerged as state-of-the-art generative models~\cite{diffusion_beat_gans}, achieving remarkable results in tasks such as image synthesis. However, training diffusion models, like other deep learning models requires large datasets, which is often difficult to obtain in biological applications.

In this work, we propose \textbf{Phen-LDiff} a method to detect cellular variations in small biological datasets by leveraging pre-trained Latent Diffusion Models (LDMs)~\cite{stable_diffusion}.

\section{Related Work}

\label{gen_inst}

\paragraph{Diffusion Models.} Diffusion Models (DMs)\cite{ddpm,ddim} are generative models that have recently achieved remarkable results in various tasks. DMs are latent variable models that operate through two key processes: a fixed forward process that gradually adds noise to the data and a learned backward process that denoises it, reconstructing the data distribution~\cite{ddpm,ddim}. Recently, these models have seen several advancements~\cite{ddim,diffusion_beat_gans,classifier_free,unified_diffusion}, making them state-of-the-art in image synthesis, surpassing traditional generative models like GANs~\cite{diffusion_beat_gans}. One of the notable improvements is the introduction of Latent Diffusion Models (LDMs)~\cite{stable_diffusion}, where images are first compressed into a latent space using a variational autoencoder, and then the diffusion process occurs within this compressed latent space. This approach enables more efficient scaling to higher-resolution images and accelerates training times. Additionally, LDMs incorporate a conditioning mechanism, allowing for tasks such as text-conditioned image generation, inpainting, and super-resolution.
These innovations in LDMs have facilitated their training on massive datasets~\cite{LAION_dataset}, resulting in powerful pre-trained models such as Stable Diffusion~\cite{stable_diffusion}, which have demonstrated exceptional performances in various generative tasks.

\paragraph{Identifications of Phenotypes in Biological Images.}

Identifying phenotypic variations in biological images is essential in biology and drug discovery~\cite{cellular_profiling,cellular_profiling_2}, yet it presents significant challenges. One of the key difficulties is the biological variability among cells within the same condition, which can obscure the differences between distinct conditions. Recently, generative models have been employed to cancel this natural variability in order to visualize and explain cellular phenotypes in microscopy images~\cite{Lamiable2023,unbiased_cell,bourou_1}. In~\cite{bourou_1}, cellular variations between conditions were identified through an image-to-image translation task between two classes, following methodologies similar to those in~\cite{cyclegan,pixtopix}. In Phenexplain~\cite{Lamiable2023}, a conditional StyleGAN2~\cite{stylegan2} was trained to detect cellular changes by performing translations between synthetic images within the latent space of StyleGAN2, allowing for training across multiple conditions, unlike the approach in~\cite{bourou_1}. A similar method was presented in~\cite{unbiased_cell}, but instead of utilizing the latent space of GANs, the authors proposed learning a representation space using self-supervised learning techniques~\cite{self_supervised_survey}. In~\cite{bourou_2}, conditional diffusion models were applied to identify phenotypes in real images. This approach consists of two stages: first, the source class image is inverted into a latent code, which is then used to generate an image from the target class. This method provides a powerful alternative for phenotype detection using real biological data. However, all of these models require a large number of images to be properly trained.

\paragraph{Fine-tuning Diffusion Models.}

Fine-tuning~\cite{fine_tunening_1,fine_tuning_1,fine_tunening_2,svdiff,lora}, a well-established strategy for training deep learning models on limited data, involves adapting pre-trained models. It involves adapting a pre-trained model's weights to fit a smaller dataset. Fine-tuning methods can be categorized into three main groups: adaptive methods~\cite{fine_tunening_1,fine_tuning3}, where the entire model's weights are adjusted; selective methods~\cite{selective_1,selective_2,selective_3}, where only a subset of the model's parameters are modified; and additive methods~\cite{lora,svdiff}, where additional networks are incorporated to refine the weights. These techniques have proven effective for discriminative models and have recently been extended to generative models, such as GANs, autoregressive generative models~\cite{lora}, and diffusion models~\cite{svdiff}. Fine-tuning techniques for diffusion models have gained attention, particularly due to the availability of models pre-trained on large datasets. Recently, several approaches have been proposed for fine-tuning diffusion models~\cite{svdiff,fine_tuning_1,lora}, driven by the popularity of pre-trained models like Stable Diffusion~\cite{stable_diffusion}. In~\cite{fine_tuning_1}, it was demonstrated that modifying a subset of parameters can lead to efficient fine-tuning. Low-Rank Adaptation (LoRA)\cite{lora}, a technique originally developed for fine-tuning large language models (LLMs)~\cite{LLMs}, can also be applied to diffusion models. LoRA freezes the pre-trained model's weights and learns low-rank matrices that are injected into each layer of the network. In\cite{svdiff}, the authors introduced SVDiff, a fine-tuning method for diffusion models that focuses on learning shifts in the model’s singular values.

\begin{figure}[h]
  \centering
  \begin{subfigure}{\linewidth}
    \centering
    \includegraphics[width=0.9\linewidth]{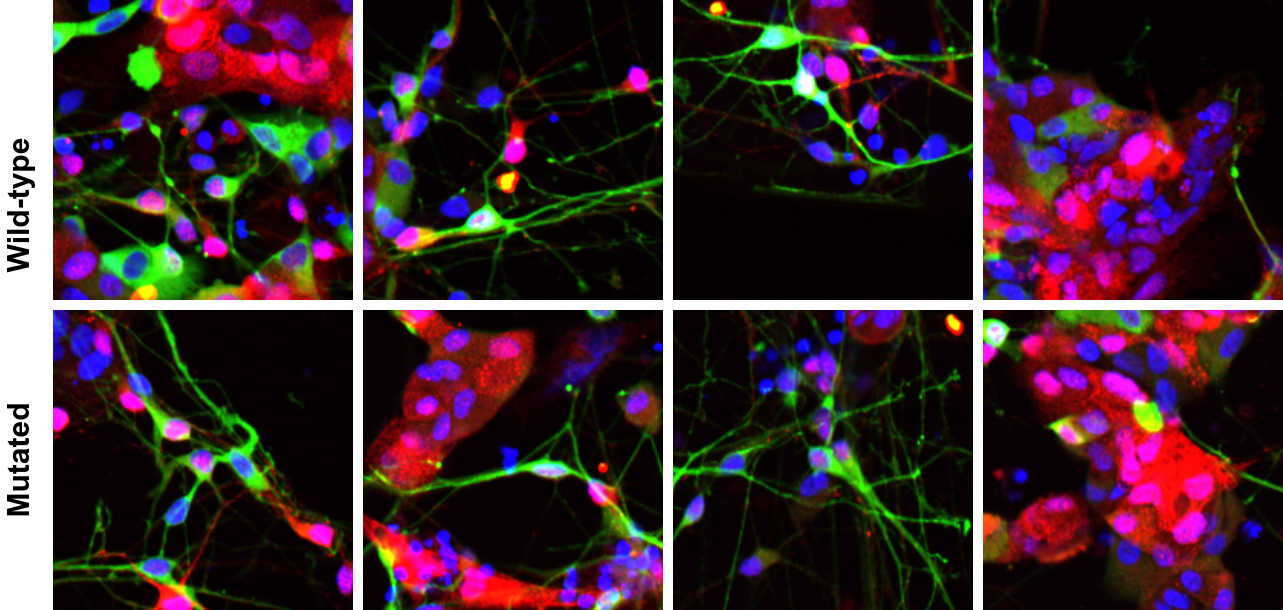} 
    \caption{}
    \label{fig:subfigure1}
  \end{subfigure}

  \vspace{0.5cm} 

  \begin{subfigure}{\linewidth}
    \centering
    \includegraphics[width=0.9\linewidth]{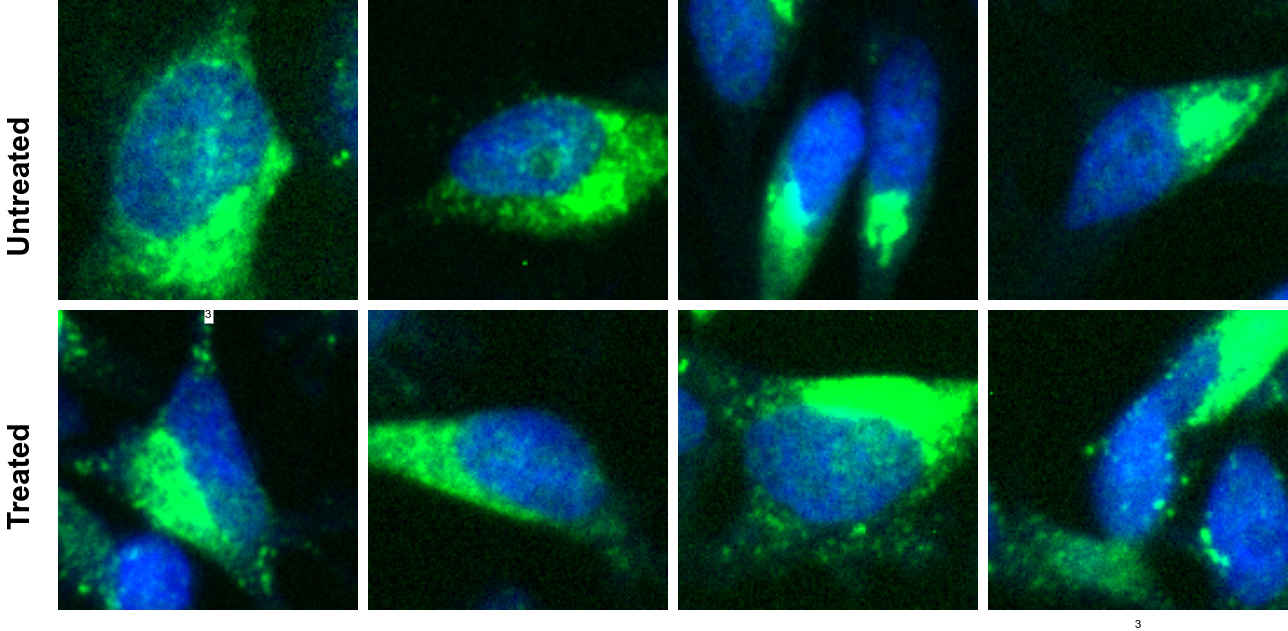} 
    \caption{}
    \label{fig:subfigure2}
  \end{subfigure}

  \caption{\textbf{Top}: Real images from the LRRK2 dataset, displaying wild-type images in the first row and images of mutated neurons in the second row. \textbf{Bottom}: Real images from the Golgi dataset, with untreated images in the first row and Nocodazole-treated images in the second row. In both (a) and (b), identifying and interpreting differences between the two classes by eye is highly challenging. However, it is essential for understanding the disease in (a) and assessing the treatment effects in (b)}
  \label{fig:example}
\end{figure}

\section{Method}
\label{sec:formatting}

\begin{figure*}
  \centering
  \includegraphics[width=1\linewidth]{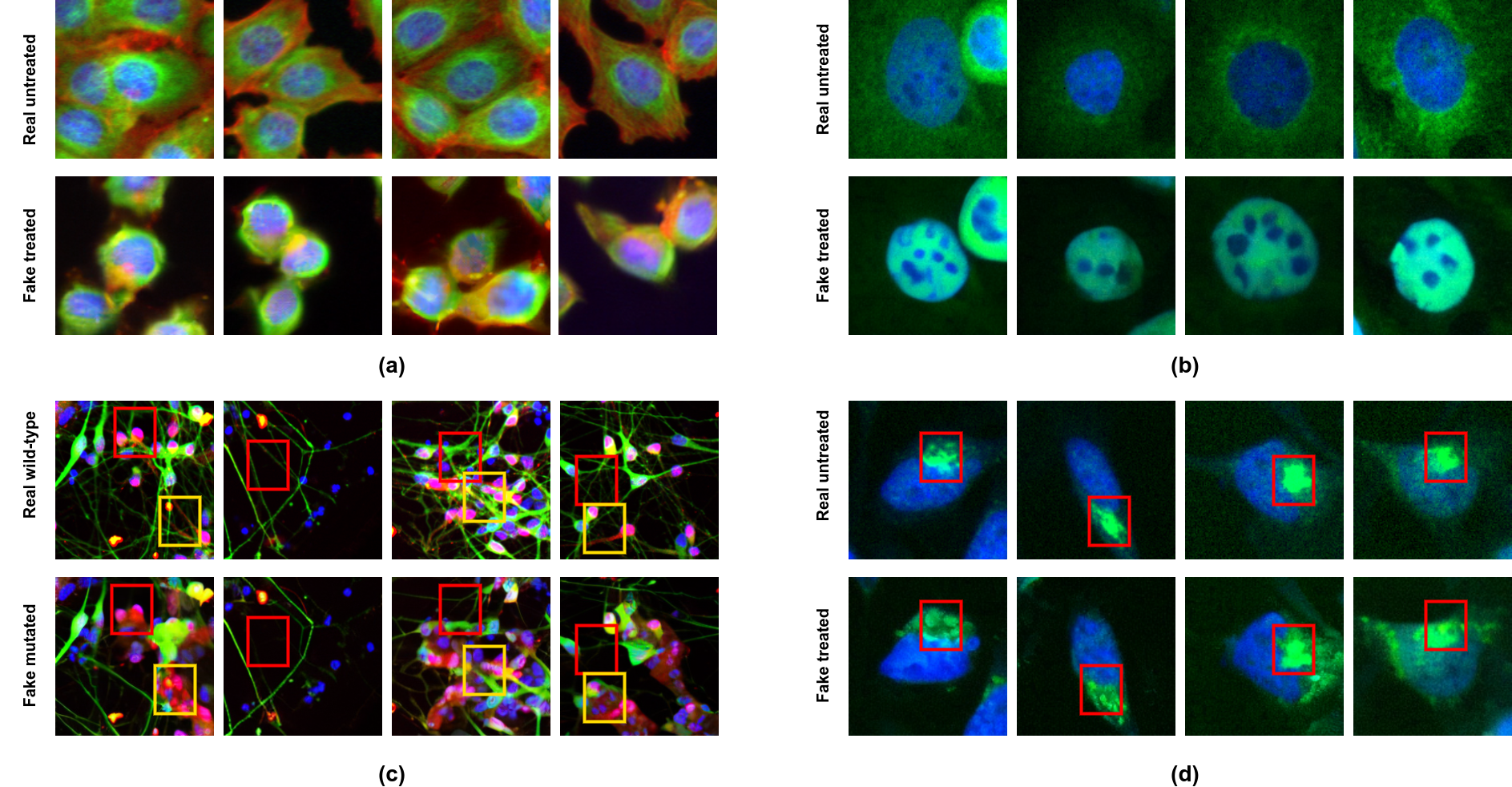} 
  \caption{We fine-tuned diffusion models on four different microscopy image datasets and performed translations from the source class to the target class. We observed the following: In \textbf{(a)}, the translated images of untreated BBBC021 samples successfully replicated the effects of Latrunculin B treatment, where we observed a decrease in cell count and the disappearance of the cytoplasmic skeleton, likely due to the toxicity of the treatment. In \textbf{(b)}, TNF treatment on cells and its translocation effect was well recapitulated by image translation. In \textbf{(c)}, we translated images of wild-type cells to images of LRRK2 mutated cells and noticed a reduction in neuron density and complexity (red squares) and an increase of $\alpha$-synuclein (yellow squares), recapitulating known effects of the mutation. Finally, in \textbf{(d)}, we observed the correct replication of the effect of Nocodazole treatment causing the scattering of the Golgi apparatus (red squares). Note how pronounced ((a), (b)) as well as subtle ((c), (d)) phenotypic changes are well captured by our model. In any case seeing the same cell before and after treatment allowed us to assess the effect of the perturbation. Real images of both conditions of the four datasets can be seen in Appendix A.1.}
  \label{fig:img_translation}
\end{figure*}

\begin{figure}
  \centering
  \includegraphics[width=1\linewidth]{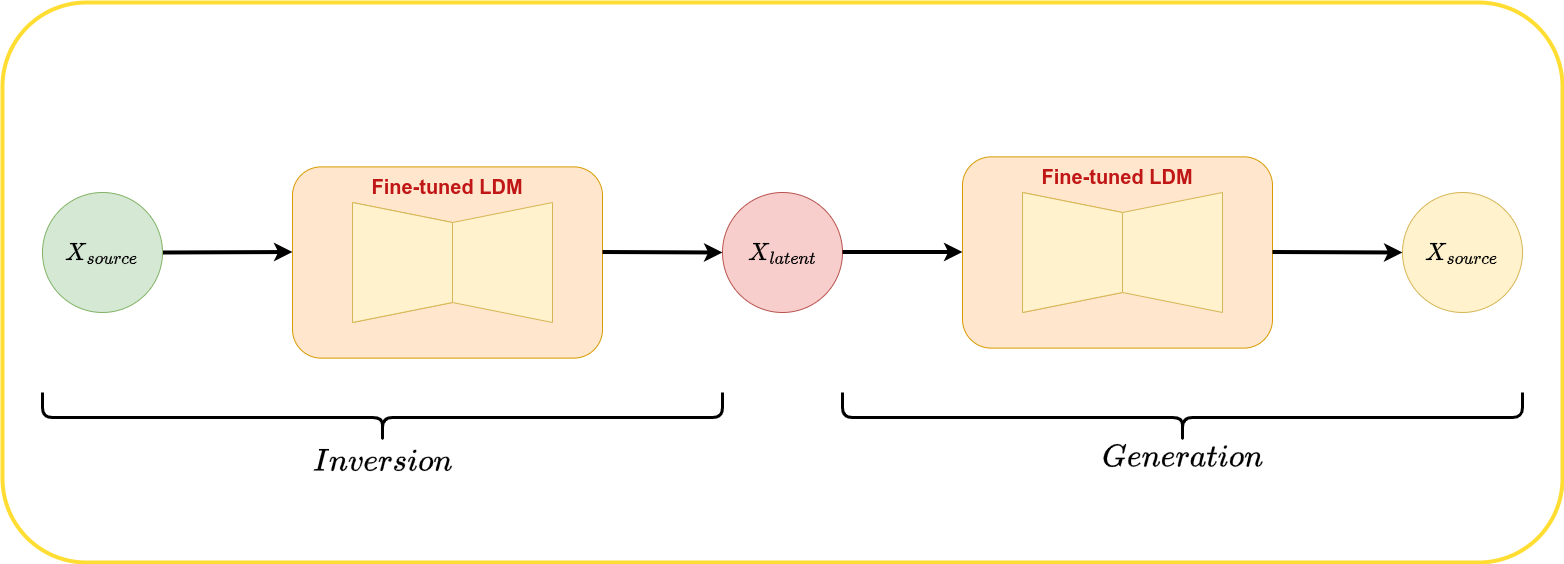} 
  \caption{Phen-LDiff leverages fine-tuned LDMs to perform image-to-image translation, identifying phenotypic variations between the images of two conditions. First, a fine-tuned model is used to invert an image from the source class into a latent code, which is then used to generate an image in the target class.}
  
  \label{fig:architecture}
\end{figure}
In this section we first provide an overview of DMs and the methods used for fine-tuning them, then we dive into the details of our approach.

\subsection{Background}
\subsubsection{Diffusion Models}  Denoising Diffusion Probabilistic Models (DDPMs) are latent variable models that utilize two Markov processes: a fixed forward process that gradually adds noise to the data, and a learned reverse process that removes noise to recover the data distribution. Formally, given data $x_0 \sim q(x_0)$, the forward process iteratively adds Gaussian noise over $T$ time steps following a forward transition kernel given by:
\begin{equation} q(x_t,|,x_{t-1}) = \mathcal{N}\left(x_t; \sqrt{1 - \beta_t} x_{t-1}, \beta_t \mathbf{I}\right) \label{eq2} 
\end{equation}
In the reverse process, noise is gradually removed using a learnable transition kernel: 
\begin{equation} p_{\theta}(x_{t-1},|,x_t) = \mathcal{N}\left(x_{t-1}; \mu_{\theta}(x_t, t), \Sigma_{\theta}(x_t, t)\right) \label{eq3} 
\end{equation}
While DDPMs generate high-quality images, they require many iterations during inference, making the process computationally intensive. To accelerate inference, \textit{Denoising Diffusion Implicit Models} (DDIMs)~\cite{ddim} can be employed. Notably, DDIMs offer deterministic sampling, allowing for \emph{exact} inversion, a property that is crucial for our approach to observe phenotypic changes in real images.

Latent Diffusion Models (LDMs)~\cite{stable_diffusion} extend DDPMs by introducing a latent space to improve both efficiency and flexibility in high-dimensional data generation tasks. Instead of operating directly in the data space, LDMs learn to encode images into a lower-dimensional latent space $\mathcal{E}(x)$, where the diffusion process occurs. This significantly reduces computational overhead, as the diffusion steps are performed on a smaller latent representation rather than on the full-resolution image. This approach not only accelerates inference but also makes it feasible to train LDMs on very large datasets.

\begin{equation}L_{LDM} = \mathbb{E}_{\mathcal{E}(x), y, \epsilon \sim \mathcal{N}(0,1), t} \left[ \left\| \epsilon - \epsilon_\theta \left( z_t, t, c \right) \right\|_2^2 \right]
\end{equation}

where: $\mathcal{E}$ is the encoder, $c$ is the condition and $\epsilon_\theta$ is the parameterized noise predictor. 

\subsubsection{Low Rank Adaptation (LoRA)}
Low-Rank Adaptation~\cite{lora} is a technique designed to efficiently fine-tune large pre-trained models by significantly reducing the number of trainable parameters. Instead of updating the entire weight matrix \( W \) during fine-tuning, LoRA introduces trainable low-rank matrices to approximate the weight updates. Specifically, the weight update \( \Delta W \) is decomposed into a product of two low-rank matrices \( B \in \mathbb{R}^{d \times r} \) and \( A \in \mathbb{R}^{k \times r} \), where \( r \ll \min(d, k) \). The adapted weight matrix during training is computed as follows:
\begin{equation}
W' = W + BA^\top
\end{equation}
This method can be either applied to all or a subset of the model layers.

\begin{figure*}

  \centering
  \includegraphics[width=\linewidth]{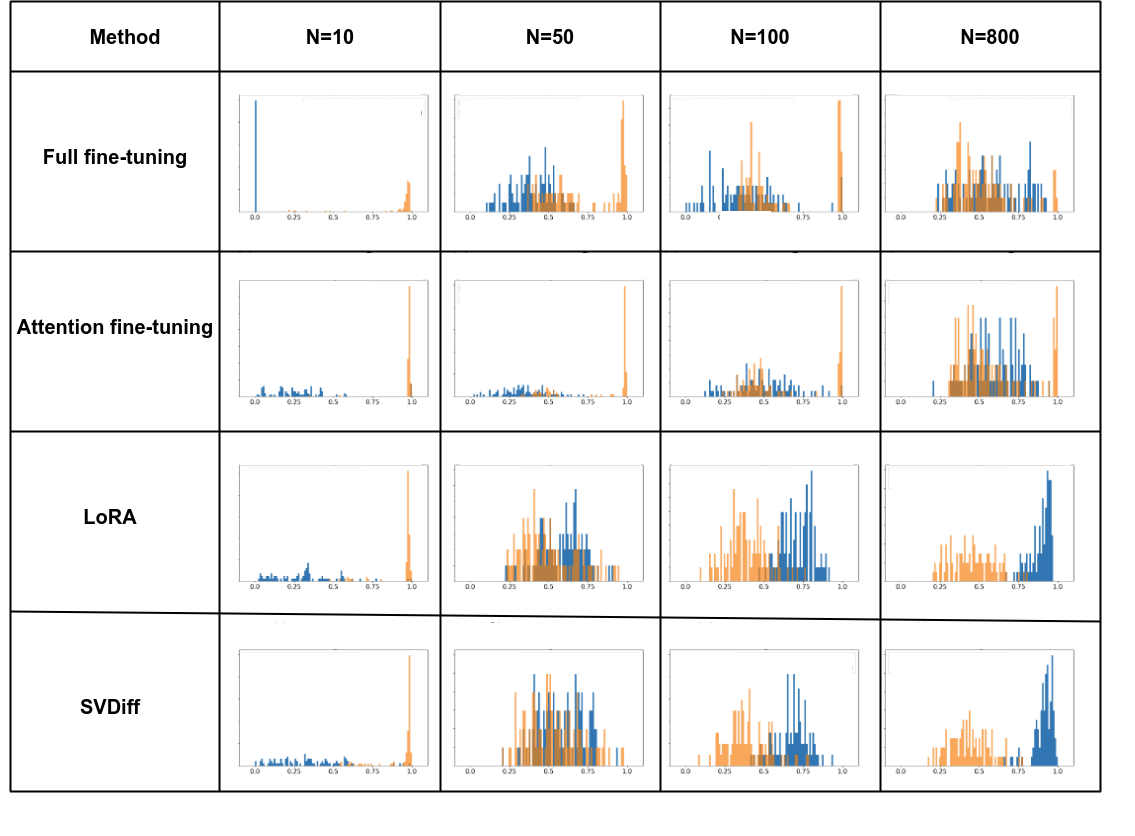} 
  \caption{Visualizing the \textbf{generalization} and \textbf{memorization} of fine-tuned diffusion models on subsets of different sizes from the BBBC021 dataset. Each plot shows two histograms: the blue histogram represents the cosine similarity between images generated using the same seed by two fine-tuned models trained on distinct, \textbf{non-overlapping} subsets of the same size. If the model has achieved generalization, the \textbf{blue} histogram should be close to one, indicating that the two images generated by the models are very similar. The orange histogram represents the cosine similarity between a generated sample and its closest image from the training dataset. A well-generalized model would produce an \textbf{orange} histogram far from one, indicating that the generated images have low similarity to any specific training example.}

  \label{fig:histograms}
\end{figure*}

\subsubsection{SVDiff}
SVDiff is a method developed to efficiently fine-tune large diffusion models by performing a \textbf{ singular value decomposition (SVD)} on the weight matrices $W$.
\[
W = U \Sigma V^\top
\]

During fine-tuning, instead of updating the entire weight matrix $W$, SVDiff updates only the singular values of this matrix. This significantly reduces the number of parameters that need to be trained, leading to faster training times and reduced computational resources. By operating in this lower-dimensional space, SVDiff helps prevent overfitting and makes it more practical to adapt large diffusion models to specific tasks or datasets.
\subsection{Datasets}
In this work, we evaluated the proposed method on several biological datasets. In some of them, cell variations are pronounced to showcase our approach, while in others, the differences are more subtle illustrating the usefulness of the method to display them. The datasets used are as follows:

\noindent \textbf{BBBC021}:
The BBBC021 dataset \cite{bbbc} is a publicly available collection of fluorescent microscopy images of MCF-7, a breast cancer cell line treated with 113 small molecules at eight different concentrations. For our research, we focused on images of untreated cells and cells treated with the highest concentration of the compound Latrunculin B. In Fig.~\ref{fig:img_translation}, the green, blue and red channels label for B-tubulin, DNA and F-actin respectively.

\noindent \textbf{Golgi}:
Fluorescent microscopy images of HeLa cells untreated (DMSO) and treated with Nocodazole. In Fig. \ref{fig:subfigure2}, the green and blue channels label for B-tubulin and DNA respectively.

\noindent \textbf{LRKK2}:
This dataset contains images of dopaminergic neurons derived from iPSCs reprogrammed from fibroblasts of a Parkinson's disease patient affected by the LRRK2-G2019S mutation. It also includes images where the mutation was genetically corrected using CRISPR-cas9, providing a rescued isogenic control~\cite{Lamiable2023}. In Fig.~\ref{fig:subfigure2} the bleu, green and red label for DNA, dopaminergic neurons and alpha-synuclein (SNCA) respectively.

\noindent \textbf{Translocation}:
Fluorescent microscopy images depicting the subcellular localization of the NF$\kappa$B (nuclear factor kappa B) protein, either untreated or treated with TNF$\alpha$ (the pro-inflammatory cytokine tumor necrosis factor alpha). In Fig~\ref{fig:img_translation} (b), the blue and green channels labels for DNA and NF$\kappa$B protein respectively.

\begin{figure}

  \centering
  \includegraphics[width=\linewidth]{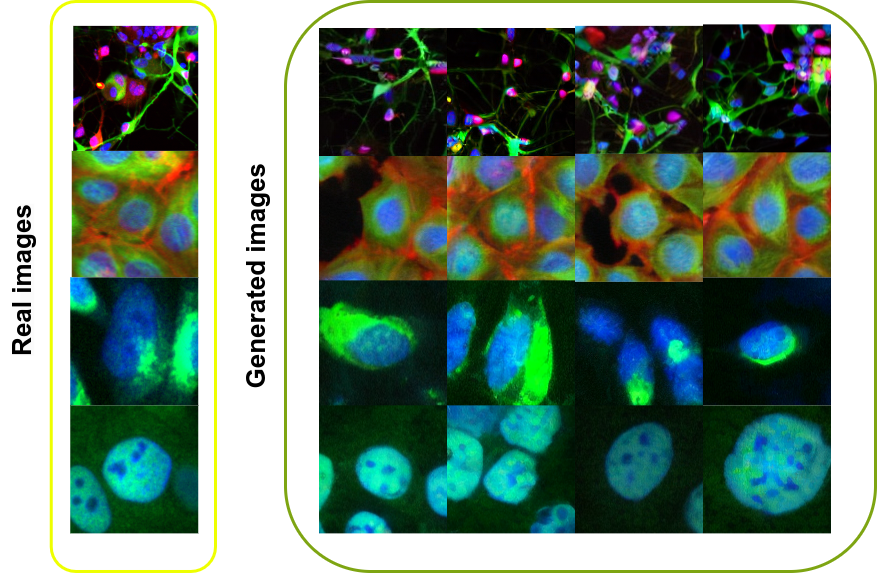} 
  \caption{ The images generated by a diffusion model fine-tuned on 100 images using LoRA on different biological datasets, we can see that the generated samples resemble the real ones. }

  \label{fig:domain_adapt}
\end{figure}

\subsection{Proposed Approach}
\begin{figure}[h]
  \centering
  \begin{subfigure}{\linewidth}
    \centering
    \includegraphics[width=\linewidth]{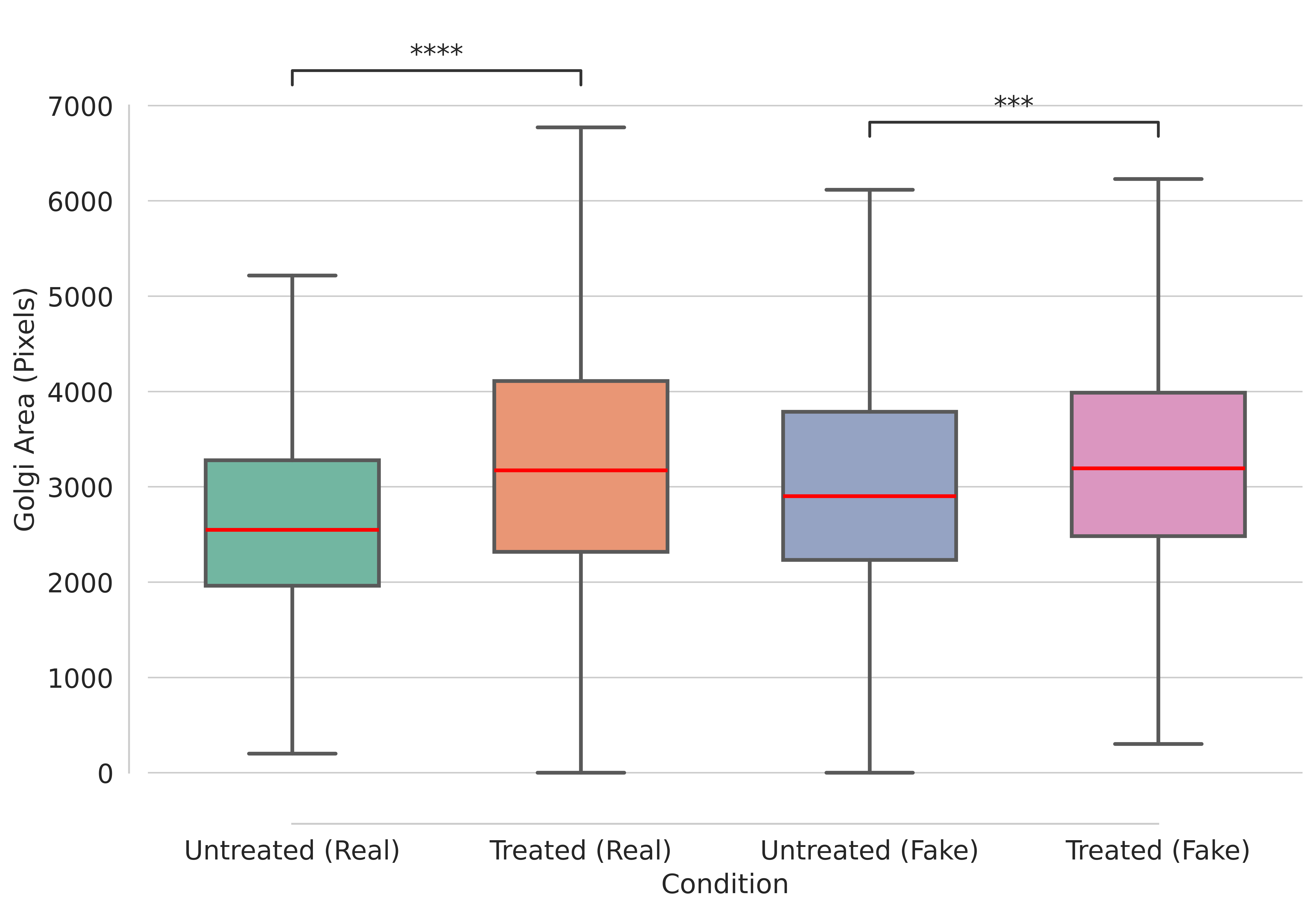} 
    \caption{The measurement of the Golgi apparatus area performed on real and synthetic images for both conditions indicates a difference in the area occupied by the Golgi apparatus, confirming the observation made by Phen-LDiff. Specifically, it appears more scattered in the treated case, which explains its larger size.}
    \label{fig:subfigure1}
  \end{subfigure}

  \vspace{0.5cm} 

  \begin{subfigure}{\linewidth}
    \centering
    \includegraphics[width=\linewidth]{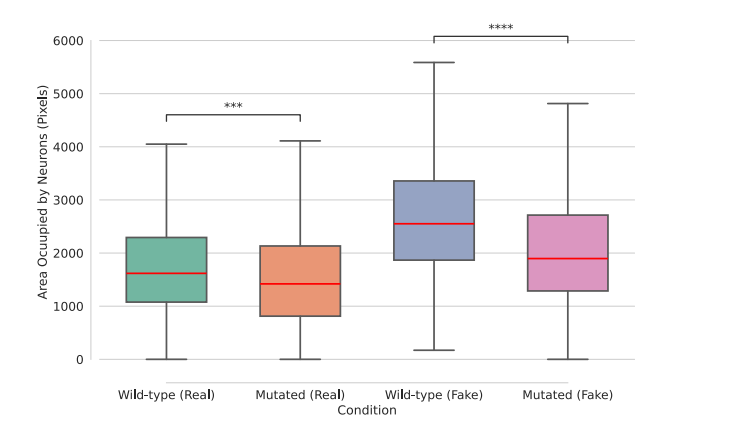} 
    \caption{The measurement of the area occupied by neurons (green channel) on real and synthetic images for both conditions indicates a reduced neuron count in the mutated case, confirming the observation made by Phen-LDiff. Indeed, the mutation that causes Parkinson's disease leads to a decrease in both the number and complexity of neurons}
    \label{fig:subfigure2}
  \end{subfigure}

  \caption{An image analysis measurement using CellProfiler~\cite{cellprofiler} on the Golgi and LRRK2 datasets, performed on real and synthetic images for both conditions, led to the same quantitative conclusions, indicating that Phen-LDiff can detect subtle cellular variations in models fine-tuned on datasets with as few as 100 images per class.}
  \label{fig:boxplots}
\end{figure}

In this work, we introduce \textbf{Phen-LDiff}, a method that leverages pre-trained Latent Diffusion Models (LDMs) for image-to-image translation on small biological datasets to identify phenotypic differences. Our approach begins by conditionally fine-tuning a general-purpose LDM on microscopy images from different experimental conditions (e.g., treated vs. untreated, wild-type vs. mutant, as illustrated in Fig.\ref{fig:example}). To perform the translation from one class to another, we first invert an image from the source class into its latent representation, which is then used to generate a corresponding image in the target class.

\section{Results}
In this work, we utilized Stable Diffusion 2, which was pre-trained on the LAION-5B dataset~\cite{LAION_dataset}. LAION-5B is a large-scale collection of web-scraped image-text pairs, encompassing a wide variety of general image sources across the internet. We fine-tuned this model on the BBBC021 dataset using several strategies: (1) full fine-tuning, where all model parameters are updated; (2) attention fine-tuning, where only the attention layers of the model are modified; and (3) LoRA and SVDiff, two techniques designed to efficiently reduce the number of trainable parameters while preserving model performance.

\subsection{Domain adaptation of fine-tuned LDMs}
As shown in Fig.~\ref{fig:domain_adapt}, the fine-tuned Stable Diffusion 2 model demonstrates the ability to generate high-quality biological images. This 
highlights the model's capability to shift its original distribution, from natural images to those closely aligned with the specific characteristics of biological data. Furthermore, the results indicate that the generated images maintain good quality across various biological datasets, even when trained on a limited number of images (100 images per dataset in our case). This suggests that pre-trained models can be effectively leveraged to learn new biological image distributions, even with a small training dataset.

\subsection{Assessing generalization and memorization in fine-tuned LDMs}
Recently, some studies have observed that diffusion models can memorize samples from the training set, leading to their replication during inference~\cite{somepalli,carlini}. This behavior was particularly noted in~\cite{kadkhodaie}, where diffusion models trained on small datasets exhibited memorization. In contrast, it was demonstrated that the same models do not exhibit this memorization when trained on sufficiently large datasets. To ensure that our fine-tuned models do not merely memorize the training datasets but instead learn the underlying distribution of the images, we adopted the approach proposed in~\cite{kadkhodaie}. Specifically, we fine-tuned two models using two non-overlapping subsets from the same datasets (thus two different samples from the same distribution) and measured the cosine similarity between images generated from the same seed, as well as the correlation between each generated image and its closest match from the training dataset. This evaluation was conducted across four different fine-tuning methods: full fine-tuning, attention fine-tuning, SVDiff, and LoRA, as illustrated in Fig.~\ref{fig:histograms}. From the results, we observe that with only 10 training images, all fine-tuning methods tend to memorize the training dataset, resulting in high correlation values between the generated images and the closest ones from the training set. Furthermore, we notice that full and attention fine-tuning struggle to generalize effectively, even as the number of training images increases. In contrast, for LoRA and SVDiff, we see that with just 50 training images, the blue and orange histograms begin to shift toward 1 and 0, respectively, indicating greater generalization and reduced memorization. Although no significant differences were observed in the quality of the generated images across the methods, we chose to use LoRA for the remaining experiments due to the more optimized and faster implementation available to us.

\begin{figure}

  \centering
  \includegraphics[width=\linewidth]{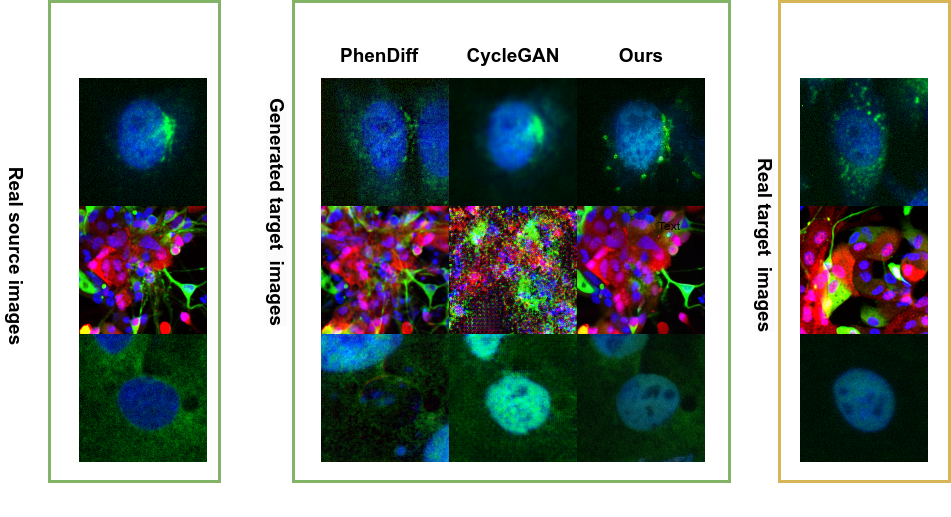} 
  \caption{We translated real untreated (Wild-type) images to the treated (mutated) condition using PhenDiff, CycleGAN, and Phen-LDiff, all the models were trained on datasets of 100 images. For PhenDiff, we can see that the translated images do not resemble the cell images in the source class but are rather new samples from the target distribution than translated cells. For CycleGAN, the translated images are very similar to the source class, but the quality is somewhat lower and the image does not recapitulate well the target class phenotype. In contrast, for the images translated with our method, we can see that they produce the desired phenotypes for the cells that were present in the provided image from the source class, indicating a successful translation.}

  \label{fig:compare:methods}
\end{figure}

\begin{figure}[h]
  \centering
  \begin{subfigure}{\linewidth}
    \centering
    \includegraphics[width=0.8\linewidth]{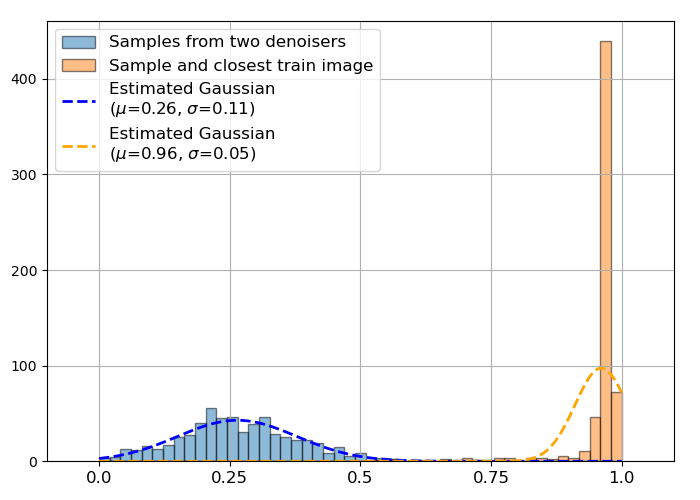} 
    \caption{}
    \label{fig:subfigure1}
  \end{subfigure}

  \vspace{0.5cm} 

  \begin{subfigure}{\linewidth}
    \centering
    \includegraphics[width=0.8\linewidth]{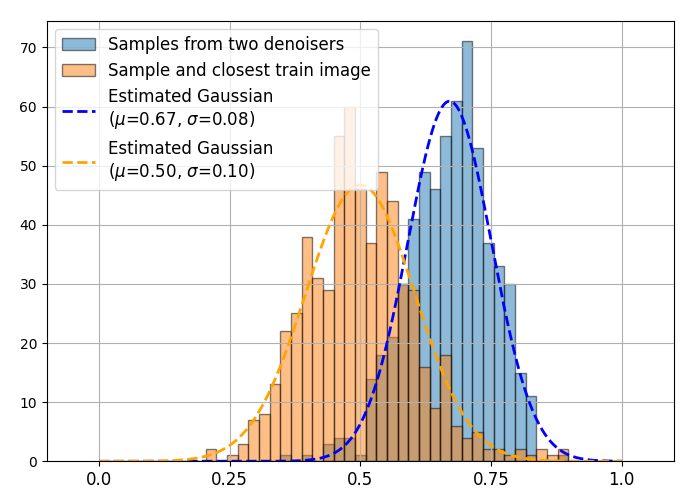} 
    \caption{}
    \label{fig:subfigure2}
  \end{subfigure}

\caption{In this figure, we trained both PhenDiff and Phen-LDiff on a subset of 50 images from the BBBC021 dataset. \textbf{Top}: The memorization histogram is close to 1, indicating very strong memorization for PhenDiff. \textbf{Bottom}: Phen-LDiff shows less memorization and achieves better generalization compared to PhenDiff.}
  \label{fig:histo_compare}
\end{figure}

\subsection{Identifying subtle cellular variations with image-to-image translation}

So far, we have demonstrated that fine-tuning Latent Diffusion Models (LDMs) is feasible even on limited biological datasets. However, our primary goal is to detect subtle cellular variations in biological samples. In Fig.~\ref{fig:img_translation}, we illustrate the image-to-image translation performed on small datasets: 100 images per class for BBBC021, Golgi, and LRRK2, and  for translocation. In Fig.~\ref{fig:img_translation} (a) and (b), the effects of treatment are visible. Specifically, for the BBBC021 dataset Fig.~\ref{fig:img_translation} (a), the phenotypic changes induced by Latrunculin B are evident. The actin cytoskeleton (red channel) has largely disappeared and a significant decrease in cell count is observed, indicating the toxicity of the treatment. In Fig.~\ref{fig:img_translation} (b), upon treatment with TNF$\alpha$, the transcription factor translocates to the nucleus, causing the fluorescence signal to shift from the cytoplasm to the nuclear region, resulting in cells displaying brightly fluorescent green nuclei. These phenotypic changes are prominent and easily recognizable. Conversely, the second row showcases more subtle phenotypes, which may be challenging to detect, even for specialists. For instance, in Fig.\ref{fig:img_translation}(d), untreated cell images from the Golgi dataset were translated to resemble treated cells. Changes in Golgi apparatus morphology due to Nocodazole treatment are noticeable, with the apparatus fragmenting into smaller stacks. In Fig.\ref{fig:img_translation}(c), when translating rescued WT images to diseased ones, we observed a decrease in dopaminergic neurons and dendritic complexity, as well as an increase in alpha-synuclein (red channel), more examples of translations can be found in Appendix A.2. To confirm these subtle observations, we used CellProfiler~\cite{cellprofiler} to quantify the changes detected by \textbf{Phen-LDiff}. For example, to confirm that the Golgi apparatus is more scattered in the treated case, we measured the area it occupies in both conditions. Similarly, for the LRRK2 dataset, we measured the area occupied by neurons (green channel) in both synthetic and real image. In Fig.~\ref{fig:boxplots}, the measurements align with the observed changes spotted by \textbf{Phen-LDiff}. Indeed, there is a significant difference between the measurements in the treated (WT) versus treated (mutated) cases, suggesting that we are identifying meaningful changes. All these now-visible differences can assist biologists in better understanding these diseases and the effects of treatments.

\begin{table*}[htbp]
\centering
\caption{Performance Metrics Across Different Datasets to evaluate}
\begin{tabular}{lcc|cc|cc|cc}
\toprule
\multirow{2}{*}{\textbf{Method}} & \multicolumn{2}{c|}{\textbf{BBBC021}} & \multicolumn{2}{c|}{\textbf{Translocation}} & \multicolumn{2}{c|}{\textbf{LRKK2}} & \multicolumn{2}{c}{\textbf{Golgi}} \\
& \textbf{FID} & \textbf{Cycle loss} & \textbf{FID} & \textbf{Cycle loss} & \textbf{FID} & \textbf{Cycle loss} & \textbf{FID} & \textbf{Cycle loss} \\
\midrule
CycleGAN & 75.98 & \textbf{528.83} & 40.56 & \textbf{643.12} & 71.23 & \textbf{428.48} & 32.28 & \textbf{341.36} \\
Phendiff & 33.31 & 2555.38 & 60.65 & 1704.54 & 74.23 & 2633.73 & \textbf{23.66} & 958 \\
Ours & \textbf{24.30} & 1707.38 & \textbf{32.79} & 1021 & \textbf{18.57} & 923.98 & 30.31 & 773.26 \\
\bottomrule
\label{tab:FID}
\end{tabular}
\end{table*}

\subsection{Comparing our method to the existing ones}
Using generative models to identify cellular variations is a growing area of research due to their potential in advancing biological studies~\cite{Lamiable2023, bourou_1, bourou_2}. Although methods like PhenExplain~\cite{Lamiable2023} can identify these variations in synthetic images, they struggle with real images due to the difficulty of inverting images using GANs. This challenge was overcome in PhenDiff~\cite{bourou_2} by leveraging the inversion properties of DDIM. However it still necessitated large datasets that are hard to get in biology. Our approach proposes the use of a pretrained latent diffusion model to enable effective performance even with limited data availability.

We compare our method to two representative models: PhenDiff, which uses diffusion models (DMs) trained from scratch, and CycleGAN~\cite{cyclegan}, which is based on GANs. As shown in Fig.~\ref{fig:compare:methods}, our method effectively highlights phenotypic cellular changes induced by the target conditions. Specifically, the Golgi apparatus appears more scattered, there is an increase in $\alpha$-synuclein, and the transcription factor translocates to the nucleus in the translocation datasets. These observations are less apparent with PhenDiff and CycleGAN. For instance, in CycleGAN, the translation quality is lower, likely due to limited data, which makes learning the target distribution challenging. In the case of PhenDiff, although some phenotypic variations are reconstructed, the translated images differ substantially from the original ones, making direct comparison with real images difficult. Additional translation examples are provided in the Appendix B.1. 

To quantitatively compare the performance of each translation method, we evaluated the quality of the translated images using FID~\cite{fid} and assessed similarity to the original images using cycle loss. For the cycle loss, an image is translated from the original domain to the target domain and back, and we compute the $L_2$ norm between the original and reconstructed images. As shown in Table~\ref{tab:FID}, our method achieves a better FID score on almost all datasets. However, CycleGAN shows a lower cycle consistency loss while producing lower-quality translations compared to the other models. This is primarily due to the cycle consistency loss used in CycleGAN training, which helps in reconstructing images but fails to produce accurate translation and thus identify phenotypic changes. Our method offers the best trade-off between capturing phenotypic variations and maintaining proximity to the initial target distribution. 

To better understand the good translation performance of our method, we compared the memorization and generalization abilities of PhenDiff and our model on 50 images per class from the BBBC021 dataset. Following the same strategy as previously described, generalization was assessed by calculating the cosine similarity between images generated from the same seed by two models trained on two independent datasets of 50 images each. Memorization was evaluated by calculating the cosine similarity between a generated image and its closest match from the training dataset. In Fig.~\ref{fig:histo_compare}, we can clearly see that PhenDiff falls into a memorization regime, whereas Phen-LDiff shows less memorization and greater generalization. Further comparisons using other datasets and sizes are presented in Appendix B.2. These results suggest that fine-tuned models achieve better generalization in low-data regimes, which explains the good translation performance of our method.

Additionally, we compared the training time of PhenDiff and Phen-LDiff on two NVIDIA L40S GPUs using the BBBC021 dataset. Training took approximately 6 hours for PhenDiff and around 2 hours for Phen-LDiff. This difference would be even more significant with larger training images, demonstrating the computational efficiency of Phen-LDiff.


\section{Conclusion}

In this work, we propose \textbf{Phen-LDiff}, a method for image-to-image translation using fine-tuned Latent Diffusion Models (LDMs) to identify phenotypic variations from limited microscopy data. Our approach demonstrates that LDMs can be effectively fine-tuned on biological datasets, capturing their underlying distributions even when data is limited. We found that certain fine-tuning approaches, such as full model fine-tuning and attention fine-tuning, can lead to memorization. In contrast, methods like LoRA and SVDiff promote better generalization, even with small datasets containing as few as 100 images per class. Our method enables image-to-image translation by first inverting an image into a latent space, followed by conditional generation to highlight phenotypic variations between conditions. We tested this approach across multiple biological datasets, showing its capability to reveal both apparent and subtle differences between experimental conditions. When compared to other representative methods, Phen-LDiff outperformed them in translation quality, even with limited image datasets.  Furthermore, our method avoids memorization and is computationally more efficient than diffusion models trained from scratch, reducing training time significantly without compromising quality. 

We anticipate that Phen-LDiff can contribute to biological research and drug discovery by enabling experts to gain deeper insights into disease mechanisms and treatment effects, especially in low-data regimes where traditional methods struggle. This efficiency and ability to generalize make Phen-LDiff a promising tool for advancing precision in phenotypic analysis.

{
    \small
    \bibliographystyle{ieeenat_fullname}
    \bibliography{main}
}


\end{document}